# Automatic Speech Summarisation: A Scoping Review

Dana Rezazadegan[1,2,*], Shlomo Berkovsky[2], Juan C. Quiroz[3,2], A. Baki Kocaballi[4,2], Ying Wang[2], Liliana Laranjo[5,2], Enrico Coiera[2]

[1]Department of Computer Science and Software Eng, Swinburne University of Technology, VIC, Australia
[2]Australian Institute of Health Innovation, Macquarie University, NSW, Australia
[3]Centre for Big Data Research in Health, University of New South Wales, NSW, Australia
[4]School of Computer Science, University of Technology Sydney, NSW, Australia
[5]Westmead Applied Research Centre, University of Sydney, NSW, Australia
[*]drezazadegan@swin.edu.au

**ABSTRACT**

Speech summarisation techniques take human speech as input and then output an abridged version as text or speech. Speech summarisation has applications in many domains from information technology to health care, for example improving speech archives or reducing clinical documentation burden. This scoping review maps the speech summarisation literature, with no restrictions on time frame, language summarised, research method, or paper type. We reviewed a total of 110 papers out of a set of 153 found through a literature search and extracted speech features used, methods, scope, and training corpora. Most studies employ one of four speech summarisation architectures: (1) Sentence extraction and compaction; (2) Feature extraction and classification or rank-based sentence selection; (3) Sentence compression and compression summarisation; and (4) Language modelling. We also discuss the strengths and weaknesses of these different methods and speech features. Overall, supervised methods (e.g. Hidden Markov support vector machines, Ranking support vector machines, Conditional random fields) performed better than unsupervised methods. As supervised methods require manually annotated training data which can be costly, there was more interest in unsupervised methods. Recent research into unsupervised methods focusses on extending language modelling, for example by combining Uni-gram modelling with deep neural networks.

Protocol registration: The protocol for this scoping review is registered at https://osf.io.

## 1 Introduction

Speech summarisation seeks to identify the most important content within human speech and then to generate a condensed form, suitable for the needs of a given task. Summarised speech should also be more understandable than a direct transcript of speech, as it excludes the breaks and irregularities, as well as the repairs or repetitions that are common in speech [2]. The steady improvement in automatic speech recognition accuracy, audio capture quality, and the increased popularity of natural language as a computer interface has underpinned the recent growth in interest for speech summarisation methods.

Speech summarisation has been applied in various settings such as broadcast news, meetings, lectures, TED talks, conversations, and interviews [3-6]. The benefits of speech summarisation range from improved efficiency and cost reduction in telephone contact centres (e.g. by identifying call topics, automatic user satisfaction evaluation, and efficiency monitoring of agents) [7] to more efficient progress tracking in project meetings [8, 9] and facilitation of learning using online courses [10, 11]. In healthcare, speech summarisation has the potential to create a new generation of digital scribes (systems which generate clinical records from spoken speech) and conversational agents which can interact with patients [12-14].

A speech summarisation system takes speech as its input and generates a summary as its output (Fig. 1).

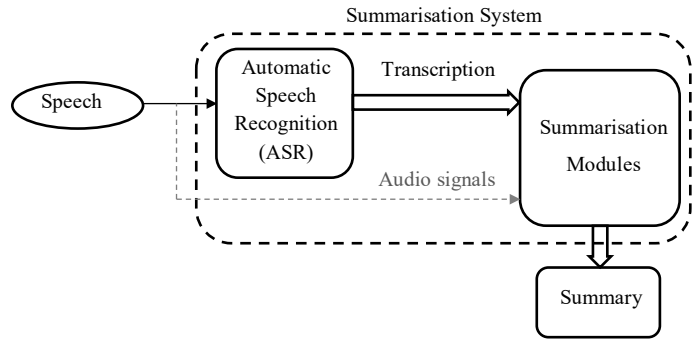

**Fig. 1: General structure of speech summarisation systems.**

| |
|---|
| **Original Text:** "lagos, nigeria (cnn) a day after winning nigeria's presidency, *muhammadu buhari* told cnn's christiane amanpour that he plans to aggressively fight corruption that has long plagued nigeria and go after the root of the nation's unrest. *buhari* said he'll "rapidly give attention" to curbing violence in the northeast part of nigeria, where the terrorist group boko haram operates. by cooperating with neighboring nations chad, cameroon and niger, he said his administration is confident it will be able to thwart criminals and others contributing to nigeria's instability." **[1]** |
| **Abstractive summary:** "*muhammadu buhari* says he plans to aggressively fight corruption that has long plagued nigeria. he says his administration is confident it will be able to thwart criminals." **[1]** |
| **Extractive summary:** "*muhammadu buhari* told cnn's christiane amanpour that he plans to aggressively fight corruption that has long plagued Nigeria. by cooperating with neighboring nations chad, cameroon and niger, he said his administration is confident it will be able to thwart criminals and others contributing to nigeria's instability." **[1]** |

**Fig. 2: Text summaries may either be extractive (using key sentences verbatim) or abstractive (inferring the meaning of text). The original text and abstractive summary were selected from [1], while we highlighted the extractive summary to show the difference.**

Speech summarisation usually involves a series of different technical components. An Automatic Speech Recognition (ASR) component first generates a direct transcription from audio into text. Next, summarisation modules (which may include sentence segmentation, sentence extraction, and/or sentence compaction sub-modules) summarise key parts of the transcription. Some summarisers skip the transcription stage with speech signal going directly into the summarisation modules [15-17].

The two approaches to speech summarisation are extractive and abstractive summarisation (Fig. 2). Extractive summarisation identifies the most relevant utterances or sentences from speech or a document that succinctly describe the main theme [18-20]. It concatenates these into a coherent summary with or without applying compression. Abstractive summarisation attempts to generate a fluent and concise summary, paraphrasing the intent, but not necessarily the exact content, of the original [21-23]. Abstractive summarisation is more challenging than extractive summarisation because of the need to infer semantic intent, as well as the need for natural language generation [21, 23].

Existing literature reviews of speech summarisation methods are now more than a decade old [24, 25] and do not include recent technical advances in deep learning, machine learning, natural language processing (NLP), and speech and text generation. For example, Furui et al. considered a small sample of papers published before 2006 and only reported on methods for extraction of important sentences and word-based sentence compaction [24]. This review aims to synthesise the existing literature on speech summarisation, including the impact of newer methods, and focuses on application domains and the speech features and training corpora used.

## 2 Search Methods

This scoping review follows the methodology outlined by Peters et al [26]. The databases searched included IEEE, Springer, Science Direct, ACM, and PubMed. We searched for papers with the keywords "speech summarisation", "conversation summarisation" and "meeting summarisation" in the title and abstract, using both British and American spelling.

Primary research studies focusing on summarising individual or multi-party speech such as lectures, news, meetings, and dialogues, were included. Research papers could be published any time up to July 2018 and be written in any language. Those papers which focused solely on speech recognition, speech analysis without summarisation, and summarisation of emails or written social media conversations were excluded. We also excluded studies. After duplicates removal, papers were screened by title and abstract against the inclusion criteria. Where information in the title and abstract was not sufficient to reach a decision, full-text screening was conducted. Fig. 3 shows the PRISMA flow diagram, with an initial set of 188 papers reducing to 110 studies which met our criteria [27]. Data was extracted from these included studies by four researchers using a standard data extraction form (Electronic Supplements, A.4). From each paper we extracted the first author's name, year of publication, title, application domain and task, training datasets (corpora), methods of summarisation, speech features, evaluation metrics, study results and key findings.

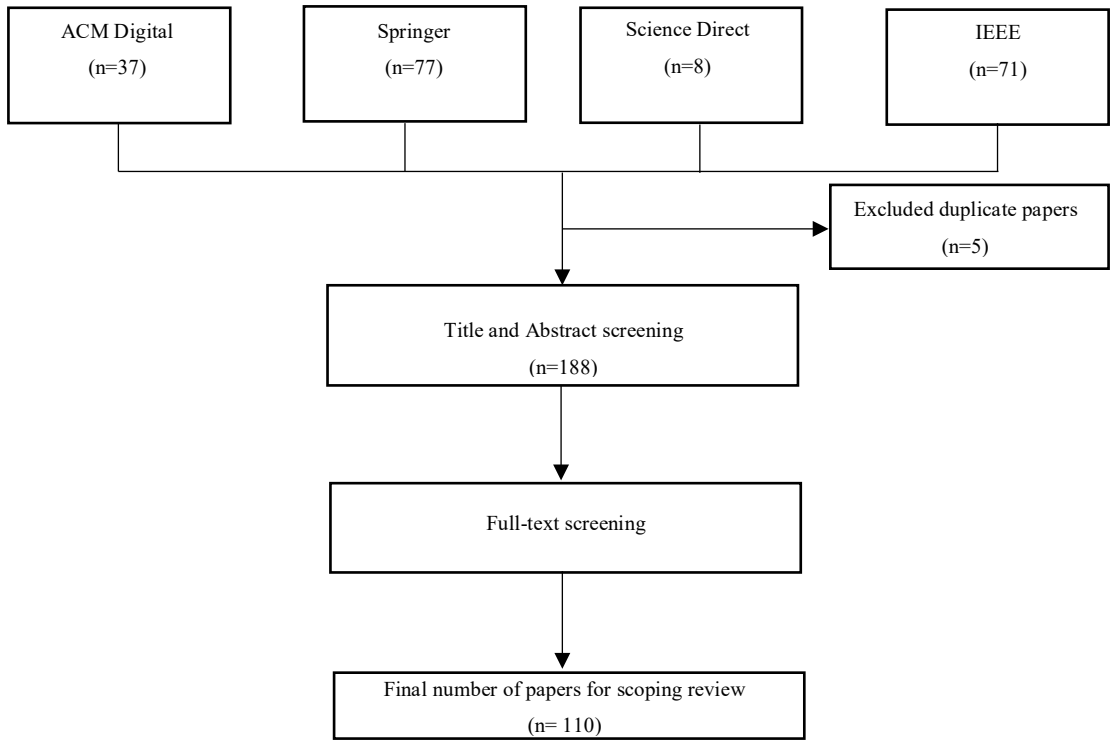

**Fig. 3: Literature search results in PRISMA format.**

## 3 Results

There was significant heterogeneity across the summarised content, application domain, technical architecture, methods, and evaluation metrics. The articles were also heterogenous in the choice of methods and speech features used, but concentrated on broadcast news, lectures and talks, meetings, interviews, and spoken conversations. This important limitation makes quantitative meta-analysis of performance results impossible and only permits characterisation of algorithms and architectures.

### 3.1 Application Domain

Most studies focused on summarising broadcast news (43 studies) and meetings (41 studies) (Table 1). This may be due to the wide availability of public, labelled datasets for broadcast news and meetings compared to the other domains where dataset availability is lower (Table 2, see Electronic supplements, A.1 for descriptions). Seventeen papers focused on summarising lectures and TED talks. Although only nine studies applied speech summarisation to conversations and interviews, the interest in this application class seems to have increased in recent years. Table 2 shows the corpora used in the reviewed papers, which are publicly available or can be provided upon request. Characteristics of available corpora in the speech domain include: Corpora mostly (1) are in English, (2) have one or two speakers, (3) are of a small/moderate size and only three go over 500 hours (See Electronic Supplements, A.1 for more details).

**Table 1 Frequency of study domains for speech summarisation studies**

| Scope/Year interval | 2002-2006 | 2007-2010 | 2011-2014 | 2015-2018 | Total |
|---|---|---|---|---|---|
| Broadcast news | 12 | 7 | 12 | 12 | 43 |
| Lectures/TED talks | 4 | 6 | 6 | 1 | 17 |
| Meetings | 2 | 23 | 8 | 8 | 41 |
| Conversation/interview | 1 | 1 | 3 | 4 | 9 |
| Total | 19 | 37 | 29 | 25 | 110 |

**Table 2 Publicly Available Corpora for Speech Summarisation Studies**

| Corpus | Summarised Content | Language | No. of speakers | Size | Source |
|---|---|---|---|---|---|
| AMI | Meeting | English | More than 2 | 100 hours | [28] |
| ICSI | Meeting | English | More than 2 | 70 hours | [29] |
| MATRICS | Multimodal meeting | English | More than 2 | 10 hours | [30] |
| TEDe | Lecture | English | 1 | 50 hours+75 hours | [31]+[32] |
| CSJ | Lecture; Task-oriented dialogue | Japanese | 1 or 2 | 658 hours | [33] |
| TDT2 | Broadcast news | English | 1 or 2 | 518-1036 hours | [34] |
| RT-03 MDE | Broadcast news+ Telephone Speech (a portion of switchboard) | English | 1 or 2 | 20 hours+40 hours | [35]+[36] |
| MATBN | Broadcast news | Mandarin | 1or 2 | 198 hours | [37] |
| ALERT | Broadcast news | Portuguese | 1or 2 | 300 hours | [38] |
| Switchboard-1 | Telephone Speech | English | 2 | 260 hours | [36] |
| Fisher | Telephone Speech | English | 2 | 2000 hours | [39] |
| MAMI | Spoken word | English | 1 | - | [40] |
| BEA | Interview | Hungarian | 2 | 250 hours | [41] |

## 3.2 Speech Features

Speech analysis can take advantage of a number of different features within an audio signal, and studies varied widely in the features used. Eight feature classes were identified (Table 3) with lexical, acoustic, and structural features most commonly used (Fig. 4).

Most studies did not provide a clear comparison of different features types, and there was little consensus on which features were most useful for summarisation. However, studies that used more than two feature types typically concluded that the best results could be obtained by combining different features [5, 42-46]. Some studies claimed that the use of only one or two features was sufficient, and performed on par with the combined use of lexical, acoustic, structural, and relevance features [10, 15, 44, 47, 48].

The importance of features differed by task and domain. For broadcast news summarisation, lexical, acoustic, and structural features achieved the best performance [20, 43-45]. Although, Chen et al. found these features to be complementary and reported the best results in combination [20], later on, they showed that using relevance features in isolation outperformed the combination of structural, lexical, and acoustic features [46]. Lin et al. also found relevance features effective in summarisation performance, whilst using the relevance features combined with lexical and acoustic features achieved the highest performance [49].

For lecture summarisation, the best indicative features were relevance and discourse features that appeared as rhetorical units (underlying message in lecture speech and corresponding slides) [5, 11, 50, 51]. These papers argued that the strong performance of relevance and discourse features is due to their resilience to problems like synonyms and recognition errors. Speaker-normalised acoustic features were found to be important when summarising lectures because of the different speaking styles of speakers [47, 52, 53].

For meeting summarisation, lexical features were identified as the most useful feature set in six papers [18, 42, 54-57]. Five papers found Term Frequency - Inverse Document Frequency (TF-IDF) to be a highly effective lexical feature [6, 18, 58-60]. Two other papers introduced the Speaker use - Inverse Document Frequency (SU-IDF) as a feature, where term weighting considers how variable is term usage across speakers in multi-speaker dialogue, and claimed that it works competitively with or better than TF-IDF [55, 61]. The combination of lexical, structural, and acoustic features was also found to achieve the highest score of Recall-Oriented Understudy for Gisting (ROUGE) in [9, 56] (See Electronic Supplements, A.3. for definition of ROUGE as the most popular evaluation metric for summarisation).

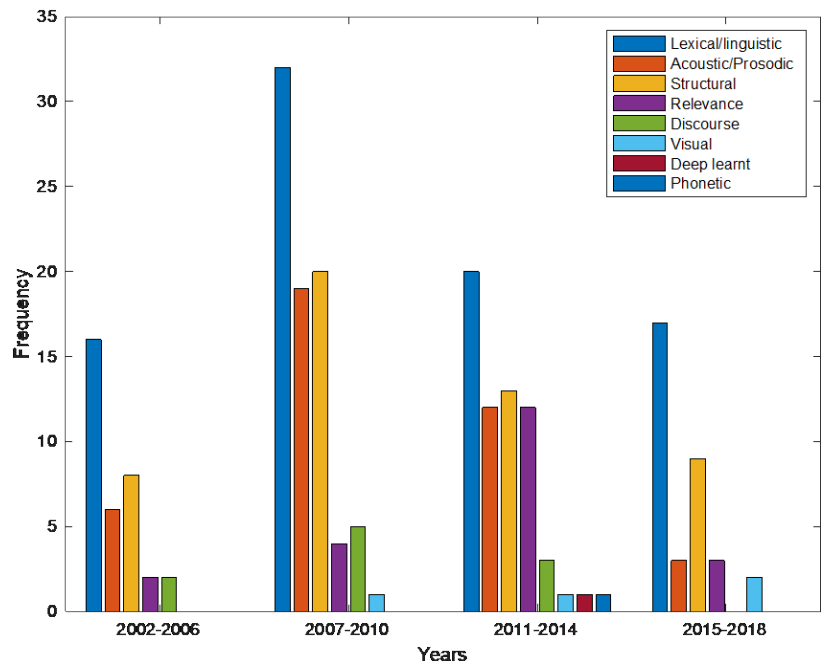

**Fig. 4: Distribution of feature categories over time. There can be studies using more than one feature type.**

**Table 3 Summary of used features in speech summarisation systems**

| Features | Definition | Examples | Frequency |
|---|---|---|---|
| Lexical (also called textual or linguistic in the literature) | Features extracted from the speech text, based on linguistic, lexical, syntax, and grammatical analysis, such as the words' size, type and semantic relationships. | · Number of words in current, previous, next utterance [45]<br>· Number of stop-words<br>· Number of NE, e.g. person, location, and organisation names [45]<br>· Number of NE which appear in the utterance at the first time in a story [45]<br>· Ratio of the number of unique NE to the number of all NE [45]<br>· TF<br>· TF-IDF<br>· POS<br>· Sentiment polarity<br>· Number of bi-gram<br>· Cosine similarity between the current utterance and the entire speech<br>· SU-IDF | 85 |
| Acoustic (also called prosodic or spectral in the literature) | Features extracted from the analysis of the audio, that can cover both features of the speaker (e.g. speaker's intention, emotion, mood, etc) and the utterance (e.g. the mode of the utterance: statement, question, command, duration, frequency, etc). | · Pauses between turns<br>· Pitch<br>· Intonation<br>· Accent<br>· Intensity<br>· Log energy<br>· Duration<br>· Spectral characteristics<br>· MFCC<br>· Frequency of utterance (F0)<br>· RMS slope<br>· Speaking rate<br>· The peak normalised cross-correlation of pitch | 40 |
| Structural | Features extracted from the structure of the utterances' transcriptions, including position of each utterance respect to t. | · Centroid scores<br>· Length of current, previous, next utterance<br>· Position of the turn in the overall speech | 50 |

| Features | Definition | Examples | Frequency |
|---|---|---|---|
| Relevance | Features extracted from the semantic similarity between all the utterances/document and each one of its utterances/sentences. | · Average similarity<br>· Relevance score obtained by VSM, Relevance score obtained by WTM, Relevance score obtained by LSA, Relevance score obtained by MRW [62] | 39 |
| Discourse | Features that typically target the presence or absence of critical words showing a planned course of action, such as decide, discuss, result, conclude, and/or phrases such as "we should". | · Number of new discourse cue/clause<br>· Discourse cue/clause position (first, second, other) [63]<br>· Position to the first discourse cue/clause [63] | 10 |
| Visual | Features extracted from videos that capture body language. | · Visual semantic concepts<br>· Objects<br>· Actions<br>· Visual attention (head-gaze target)<br>· Body behaviour<br>· Head motion<br>· Hand gestures | 4 |
| Deep learnt | Features learned directly from the speech text with the use of deep neural networks, without manual feature engineering. | · Language modelling<br>· Language patterns | 1 |
| Phonetic | Features extracted from the audio which is used for computing the similarity between sentences and sentence-like units (SUs), to overcome speech recognition errors and disfluencies [64]. | · Type (vowel/constant)<br>· Vowel length, height and front-ness<br>· Lip rounding<br>· Consonant type<br>· Place of articulation<br>· Consonant voicing<br>  (All were taken from [64]) | 1 |

Abbreviations: NE: Named Entities; TF: Term Frequency; TF-IDF: Term Frequency - Inverse Document Frequency; POS: Part-Of-Speech; SU-IDF: Speaker use - Inverse Document Frequency; MFCC: Mel-Frequency Cepstral Coefficients; VSM: Vector Space model; WTM: Word Topic Model; LSA; Latent Semantic Analysis; MRW: Markov Random Walks.

## 3.3 Speech Summarisation Types, Architectures, Methods, and Evaluation metrics

### 3.3.1 Speech Summarisation Types – Extractive vs Abstractive Summarisation

Extractive summarisation methods were more popular than abstractive methods (99 out of 110 papers, Fig. 5). However, the rate of abstractive summarisation usage increased 5.7% for 2007-2010 and 17.7% for 2011-2019.

Extractive summarisation has been criticised for propagating word errors to the summary [21, 63], for summaries being repetitive [65] or containing incomplete sentences [22, 63, 65], or for the creation of incoherent summaries [63, 66]. Only one paper showed abstractive summarisation performing better than extractive summarisation, when evaluated using ROUGE [66]. However, the results of this study are atypical, given that the ROUGE scores for both types of summarisation were about 50% lower than reported by others. This may be because they used "human abstracts rather than human extracts" as reference summaries for the evaluation, and because the summarisers were configured to generate very short summaries that leaded to extremely low recall [66]. Since other studies using abstractive summarisation methods have not performed a direct comparison to the extractive methods or have not used standardised evaluation metrics, no clear conclusion regarding the performance of the two types of summarisation can be drawn.

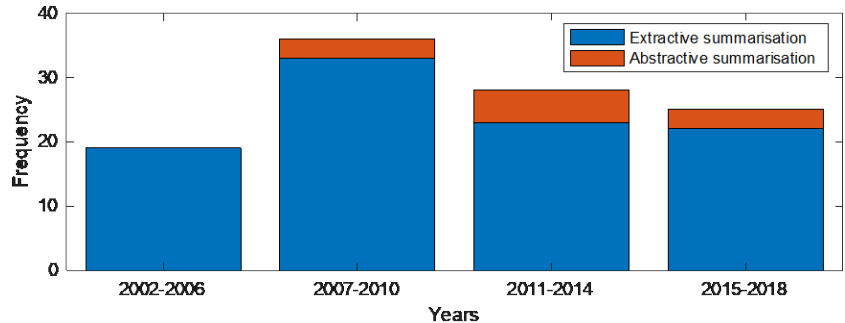

Fig. 5: Types of speech summarisation-extractive vs abstractive.

### 3.3.2 Format of output summary – Text vs Audio

Most papers (107 out of 110) generated a text summary, probably because of the convenience of using transcribed text as an intermediate representation prior to feature extraction and summarisation [24]. This is partly due to the wider selection of existing natural language processing techniques to pre-process the text and perform feature extraction [18, 42, 54-57].

Only three studies generated an audio summary, avoiding the need to have an intermediate step of transcribing speech and using text features [15, 16, 67]. These papers argued that there were advantages to preserving the original format of the data source, such as exploiting acoustic information, e.g. pauses, speaking style, and the emotion of speakers. Another reported benefit was avoiding possible speech recognition errors in the text. In general, there were two ways of audio summary generation: concatenating utterances that were extracted from the speech or generating audio from a text summary using a speech synthesiser [24]. Although the former method is likely to generate unnatural sounds which is undesirable, it is more robust against ASR errors, compared to the latter.

### 3.3.3 Speech summarisation architectures

Various summarisation architectures have been applied to the speech summarisation problem with four major approaches (Fig. 6): (1) sentence extraction and compaction, (2) binary classification of sentences or rank-based sentence selection, (3) sentence compression and summarisation, and (4) language modelling.
8

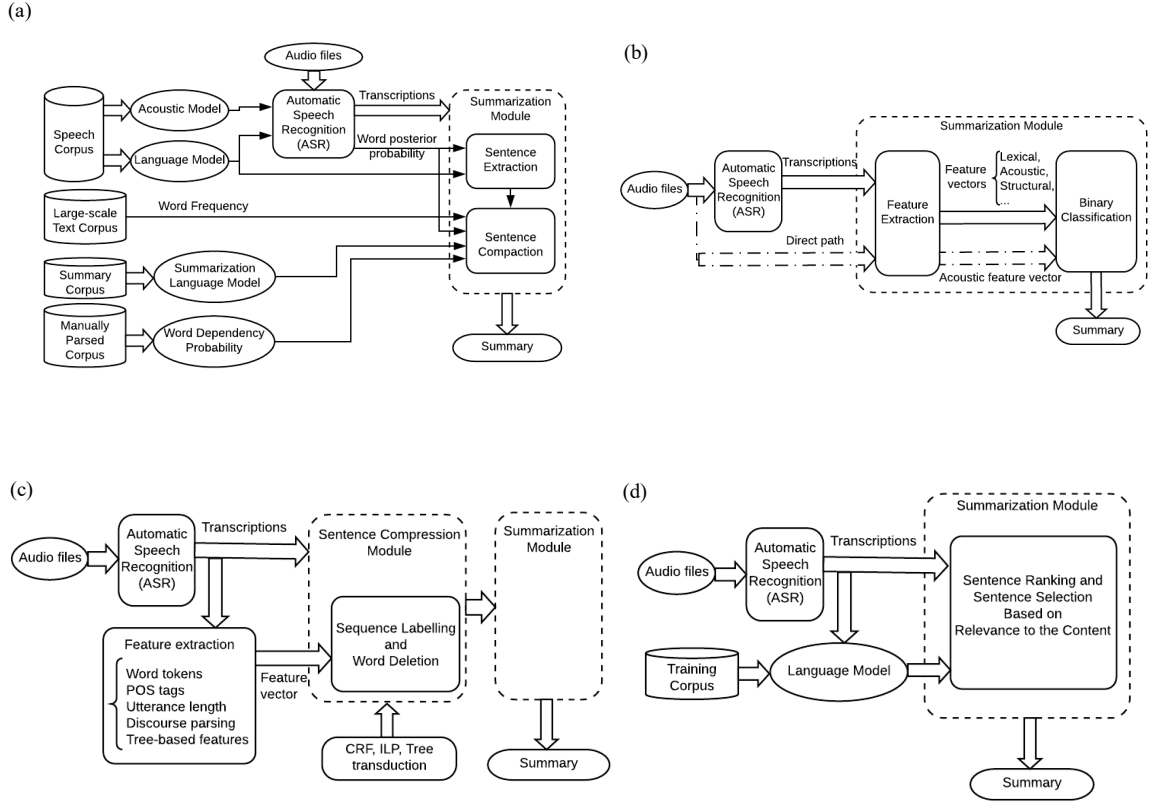

**Fig. 6** Four different architecture for speech summarisation: (a) Two-stage methodology based on sentence extraction and compaction [68]; (b) Feature extraction and binary classification; (c) Two-stage methodology based on sentence compression and summarisation; (d) The summarisation task based on language modelling

We define several technical expressions used in the following sub-sections, in Table 4.

**Table 4 Definitions of technical terms used in different architectures**

| Technical term | Definition |
| --- | --- |
| Filler phrases (FPs) | Words that can be removed without losing any information, including "discourse markers (e.g., "I mean", "you know"), editing terms" [65], and commonly used terms (e.g. "for example", "of course," and "sort of") [65]. |
| Summarisation ratio | "The ratio of the number of words in the automatic summary to that in the original transcript of a spoken document" [69]. |
| Summarisation scores | Numerical scores, including "Linguistic score, significance score and confidence score", that indicate the appropriateness of a summarised sentence through the ranking process. Please see [70] for details. |
| Sentence ranking | A ranking function which puts different weights on the summarisation scores, and then combines them. |

### 3.3.3.1 Sentence extraction and compaction

Sentence extraction and compaction (Fig. 6-a) was most popular between 2002 to 2006 and was used in six papers [69-74]. Using output generated by a speech recogniser, each sentence is scored using linguistic, significance, and confidence scores, with filler phrases removed. Then, sentence compaction is enforced to the most highly ranked sentences using dynamic programming. The summarisation ratio is set experimentally by trading off the ratio of sentence extraction and sentence compaction to gain the highest summarisation performance. Among the reviewed papers, there have been some observations on the performance and effect of different sentence scoring methods. Hori et al. argued that the linguistic score may reduce out-of-context word extraction from human disfluencies and recognition errors [3, 72]. However, only Huang et al. reported the linguistic score as the single most effective score for the summarisation of the Mandarin language [75]. Three papers found the significance score or its combination with the confidence score more important than the linguistic score, in achieving better results [68, 70, 72], which was verified for the Japanese and English languages, but the effect of confidence score was higher for English. All the studies [3, 68, 70, 72, 75] agreed that the summarisation score could be improved by combining all three scores. Eight papers used dynamic programming for sentence compaction and finding the best summarisation results [3, 68-72, 75, 76]. Seven papers used only lexical features for sentence extraction and compaction [69-74, 77], whereas two others used both lexical and acoustic features [3, 68].

### 3.3.3.2 Binary classification of sentences or rank-based sentence selection

Fig. 6-b illustrates the process of summarisation by feature extraction and binary classification. This casts the problem of summarisation as a binary classification problem, where a classifier determines whether each sentence, phrase, or speaker turn should be extracted for the summary. This approach relies on supervised learning. Simple machine learning methods, binary support vector machines (SVM) in particular, were popular until 2009, with recent papers using them for baseline comparisons [5, 20, 43, 46, 47, 49, 57, 62, 63, 78-88]. Among the reviewed papers, the following methods were used: simple SVM [5, 42, 44, 45, 51, 53, 59], Ranking SVM [62], hidden Markov SVM (HMSVM) [11, 45, 89, 90], Bayesian classifier [91], multi-layer perceptron [92, 93], conditional random field (CRF) [11, 45, 54], deep neural networks [11], logistic regression [94], random forests [95], and decision trees [18]. Zhang and Yuan compared three classifiers and found CRF to outperform HMSVM and binary SVM [45].

Imbalanced datasets were considered an issue due to insufficient number of positive class samples (summary sentences) and data resampling was used in three papers to deal with imbalanced datasets [85-87]. Two studies handled the imbalanced data problem with Ranking SVM and AdaRank summarisers using training criteria based on evaluation metric, rather than sets of features [46, 62]. These improved the summarisation performance by maximising the correlated evaluation score or by optimising an objective function linked to the final evaluation. Eight papers studied the ranking score method (rather than binary classification) after the feature selection process. This method chooses the highest ranked sentences according to their features' scores [6, 9, 61, 68, 73, 74, 77, 96]. Key phrase extraction was applied as the initial step for summarising broadcast news [75] [97], lectures [98], spoken conversations [60], and meetings [98-101].

### 3.3.3.3 Sentence compression and summarisation

Fig. 6-c illustrates the process of summarisation starting with sentence compression followed by the final summary generation [102]. Sentence compression has been formulated as a word deletion and sequence labelling task, with the aim of retaining the most important information in the form of grammatical sentences [102] [65]. Both supervised and unsupervised methods have been used for this purpose. Two papers exploited CRFs to compress the spoken utterance through a sequence labelling task while effectively integrating acoustic features [102] [65]. A filler phrase detection module and Integer Linear Programming (ILP) have been also deployed, without the need for human annotations [65]. The sentences were first compressed and then summarised using maximum marginal relevance [102] [65]. One paper applied compression methods,

such as Integer Programming (IP) and Markovisation of Synchronous Context Free Grammar, to extractive summaries to generate abstractive summaries [23].

*3.3.3.4      Language modelling*

Language modelling (LM) has attracted more attention since 2014. LM is an approach ranking each sentence or utterance of a document, based on the individual word frequency, semantic relationship, or sentence position in a document. It can be deployed either independently or in conjunction with classification methods and autoencoders (Fig. 6-d). With unsupervised extractive summarisation, LM was used to select sentences, mostly from the transcriptions generated by ASR engines [78, 103, 104]. This approach used a probabilistic generative paradigm to rank every sentence of an utterance and build a unigram language model (ULM) in accordance to the individual word frequency in the sentence. However, it has not leveraged long-span context dependence cues that could render another proof for existing the semantic relationships among words or between a given sentence and the whole document. This is taken into account in recurrent neural network LM (RNNLM) as a promising modelling framework for speech recognition [105]. An RNNLM framework, consisting of the input, hidden and output layers, has been explored for sentence modelling formulation in LM-based summarisation approaches [106, 107].

Position-aware LM for extractive summarisation was introduced in [84]. The motivation behind this method was the assumption that a sentence at a particular position of the transcription might contain more important content, or might be resided closer to the parts of the transcription that are richer in content than the rest, and consequently these sentences are better candidates for the summary. The authors tested four types of position-based LMs and the passage-based model was found to outperform others [84]. Translation-based LM was applied to score a document based on a translation model [108]. This LM approach calculates matching degree between a word in a sentence and semantically similar words in the spoken document and scores the sentence, accordingly [108].

### 3.3.4   Methods of summarisation

Machine learning methods applicable to the summarisation task can divided into three categories: supervised (sample labels required for model training), unsupervised (no labels required), and semi-supervised (small set of sample labels required). We refer readers to Electronic Supplements, A.2 to see examples of supervised, unsupervised, and semi supervised methods, supported by methods' descriptions. Whilst a variety of modern machine learning methods were applied to speech summarisation, search found only one study that deployed deep learning [107]. The distribution of the utilised methods over time (2002-now) is shown in Fig. 7.

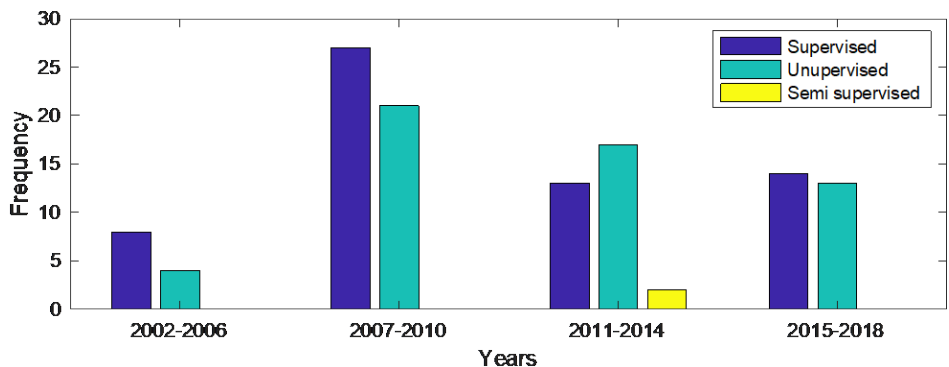

**Fig. 7 Utilised methods of speech summarisation, over time**

It shows supervised and unsupervised methods were used about the same. Supervised methods with balanced labelling of the summary and non-summary classes achieved a higher ROUGE score than unsupervised methods [46], [80]. Only two studies used semi-supervised learning [109, 110].

Directly comparing the performance of the deployed methods and frameworks was challenging, because the proposed methods were tested using different types and volumes of data with varying assumptions, restrictions, and ASR engines. Here, we only report on the main methods applied in the reviewed papers, and group them based on the domain of summarisation.

Broadcast news

Supervised methods such as AdaRank, Ranking SVM, and traditional SVM with balanced labels achieved a higher performance than unsupervised methods including MRW, VSM, WTM, LSA [46]. For instance, Kullback-Leibler divergence Measure (KLM) and SVM outperformed graph-based and vector space-based methods [81]. Among the supervised methods, CRF, HMSVM, and Ranking SVM outperformed SVM [45, 80]. Among the unsupervised methods, Uni-gram and/or Bi-gram language modelling did not beat neither the graph-based methods e.g. LexRank and MRW, nor the combinational optimisation methods such as Submodularity (SM) and ILP [106, 107]. Nevertheless, combining Recurrent Neural Network and Uni-gram language modelling techniques for extractive summarisation outperformed all the previously-mentioned methods [106, 107]. Density peaks clustering algorithm demonstrated better results than graph-based, vector-based, ULM, and word embedding-based methods [111].

Lectures

Lecture summarisation initially exploited methods based on sentence ranking [68, 73, 74, 77]. However, recently there has been a large number of papers that emphasised the usefulness of rhetorical information in a shallow or deep structure [5, 11, 47, 51, 89, 90]. HMM states and SVM classifiers have been a common trend in these papers, compared to using PLSA [50, 98] and leveraging active learning combined with SVMs [52, 109].

Meetings

In the reviewed papers that summarised meetings, the focus has mostly been on the finished actions and items/decisions as the key parts of the output summary [9, 112]. Some researchers leveraged personal notes of meeting participants as potential predictors of important meeting parts, through abstractive [113] or extractive summarisation process [18, 58], using a binary (summary, non-summary) or ternary (definitely show in the summary, maybe show, do not show) labelling. Graph-based methods achieved a better performance than vector-based methods and combinational optimisation methods [114], while the positive effect of word stemming and noise filtering for utterances has been shown to achieve a better performance on ASR transcripts than manual ones [115]. Resampling techniques have been explored and demonstrated good results, using SVM and learning-based sampling method [87].

Lately, deep neural network (DNN) approaches, i.e. CNN and fusin-3D SCNN, have been proven to be effective [116]. DNNs demonstrated a strong potential to capture short important utterances that are usually hard to extract using other methods. There have also been attempts to conduct abstractive meeting summarisation. One way was to apply sentence compression to extractive summaries, using integer programming and Markovisation of Synchronous Context Free Grammar [23]. Others focused on the combination of compression and abstractive synthesis [117], manually creating abstractive summaries [113] and using a two-stage method on ILP/CRF based sentence compression and MMR-based summarisation [65]. Another approach relied on discourse relations, in order to learn a general semantic structure and link the turns in a conversation [21].

### 3.3.5 Evaluation metrics

In this section, we report on the evaluation metrics used in studies and their strengths or limitations. A description of evaluation metrics including quantitative and qualitative metrics and the reported results are given in Electronic Supplements, A.3.

Automatically generated summaries have been evaluated using both subjective/qualitative metrics such as readability, coherence, usefulness, completeness, and objective/quantitative metrics such as ROUGE, Precision, Recall, F-measure, word accuracy, and Pyramid. Fig. 8-a shows the distribution of the qualitative and quantitative evaluation metrics over time, while the more prevalent quantitative metrics are further broken down in Fig. 8-b.

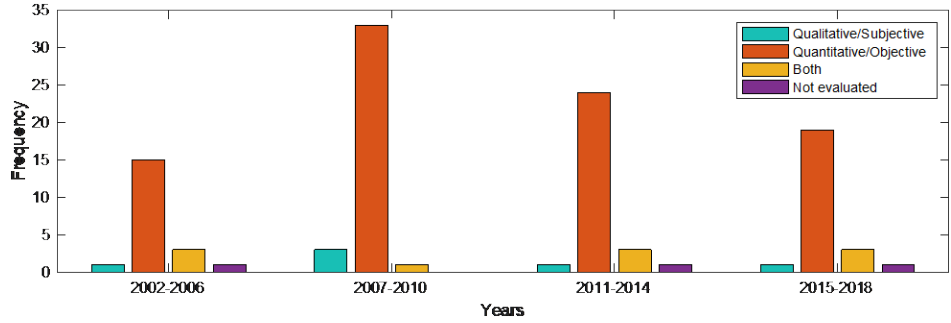

(a)

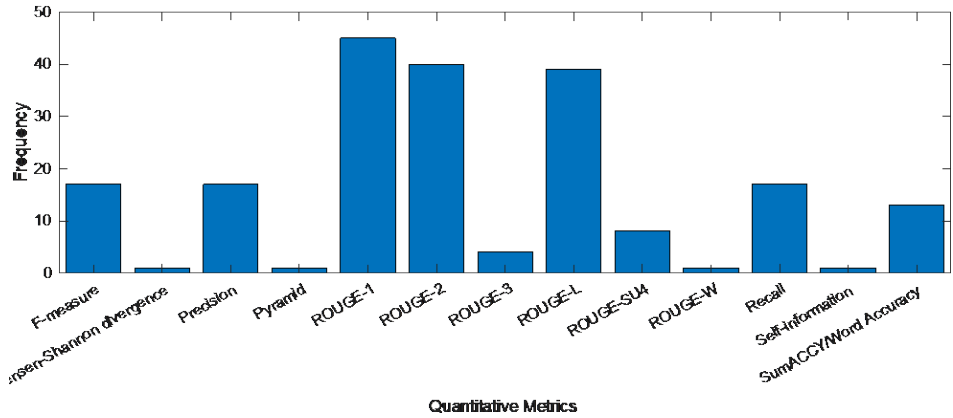

(b)

**Fig. 8 (a) Distribution of the utilised evaluation metrics; (b) Distribution of quantitative evaluation metrics**

Among the quantitative metrics, ROUGE and its variants were dominant. ROUGE has been criticised for only performing string matching between the summaries without incorporating the meaning of words or words sequences [118, 119]. Therefore, it is possible to get a high ROUGE score for a poor summary [119]. Furthermore, obtaining perfect ROUGE scores may be impossible even for humans [120]. The correlation between a ROUGE score and human evaluations was lower than expected for meeting summaries, although ROUGE-SU4 showed a better correlation than ROUGE-1 [121]. These scores could be improved by removing disfluencies and exploiting speaker information [121]. One paper reported that weighted precision, recall, and F-measure are more stringent and stable evaluation mechanisms than ROUGE for meeting summarisation [112].

Two papers recommended the use of extrinsic metrics over intrinsic ones. Intrinsic metrics evaluate summaries based on how well the information content can match the reference summaries' content [122]. Extrinsic evaluation not only aims to find the informativeness of the summary, but also to understand its usefulness in including an actual information which is required [122]. In other words, intrinsic evaluation focuses on informativeness and coherence of output summaries, whereas extrinsic evaluation targets their utility in a given application [118]. Decision audit is an example of applying this to summarise useful information related to meeting decision making [122].

One common observation amongst the reviewed studies was that human evaluation can be subjective, either as a reference summary to be used in a quantitative metric or as a direct qualitative metric. Since this affects the evaluation of automatically generated summaries, using multiple reference summaries written by different human subjects was proposed as a way to overcome this problem, although precision and recall can fail for evaluating the automatically generated summary [123]. However, this is still subjective and it requires extra analysis of the agreement among human subjects [124]. A good example of examining the consistency of human evaluation and the critical factors for human agreement was presented in [124]. Through several measurements, e.g. Kappa coefficient, ROUGE and their proposed sentence distance score and divergence, it has been shown that the number of speakers could be related to the human subjects' agreement, while the speech length was found non-critical. Another observation was the lack of agreement between subjective and objective evaluations on the performance of lexical and acoustic features, for lecture summarisation, possibly due to the relatively large number of fillers included in a lecture [68].

# 4 Discussion

## 4.1 Main findings

This scoping review identified several important gaps in the speech summarisation research literature. Although various techniques for speech summarisation have been proposed, there is still a considerable gap between the quality of automatic speech summarisation and manual summarisation by humans. Despite their potential usefulness, there has been little research on abstractive summarisation. This is partially due to the lack of suitable resources, corpora, and reference summaries in the speech domain. Another gap is the scarcity of extrinsic or task-based evaluations, which indicates that most studies focussed on traditional summarisation without paying attention to the usefulness for a specific task. The use of different corpora or different batches of the same corpus makes replication and comparison across studies difficult.

Factors such as audio quality, structured speech, and number of speakers, affect the quality of the speech-to-text conversion, selection of methods and/or features, and the overall quality of summarisation [25].

In broadcast news, the audio is recorded under ideal acoustic conditions and the speech is produced by professional following a structured script. These factors facilitate low word error rate in ASR, also making it easier to summarise [125]. Lectures are less structured, speakers are usually not trained, and speaking styles and/or accents can vary widely. Conversations and meetings tend to be even less structured, include disfluencies, interruptions, multiple speakers, and grammatically wrong utterances.

In terms of the speech features used, there was substantial variation, suggesting that the choice of feature types depends on the task, dataset, method applied, and language characteristics.

In broadcast news, language modelling is increasingly popular, but not in other domains where the data is not structured. In lecture summarisation, recent studies shifted from sentence ranking-based method [68, 73, 74, 77] to rhetorical information-based methods in a shallow or deep structure [5, 11, 47, 51, 89, 90], due to their higher performance. However, there was one common observation about the superiority of relevance features over structural, lexical, and acoustic features used in isolation [5, 11, 46, 49-51, 97]. This can be due to the nature of the "relevance features that capture the relevance of a sentence to the whole document and the relevance between sentences" [46], or can be associated with their lower vulnerability to problems like synonyms and speech recognition errors. In meeting summarisation, graph-based methods showed a better performance than other methods. Although capturing short important utterances is still challenging for these methods, DNNs showed a strong potential to address this problem [116].

## 4.2 Study Quality

There was significant heterogeneity across the reviewed articles in terms of language (English, Mandarin, Japanese, Portuguese, Persian, Hungarian, Turkish, etc.), content and size of the utilised datasets (some used their own collected data which is not publicly available, e.g. [18, 96], some used different parts of the same public dataset [73, 77]). Evaluation metrics used for assessments varied widely. This makes any quantitative meta-analysis of results impossible, and only allows the reporting of qualitative patterns across studies. This also means that there was little to no attempt by researchers to replicate earlier results.

Some researchers did not make their own collected data available, e.g. [108, 115], or did not mention how they chose a portion of data from a public corpus [73, 77]. Therefore, reproducibility was a big limitation of several papers. For example, Chatain et al. used nine talks from the TED corpus and five news from the CNN corpus, without specifying the selected talks and news stories. Chen et al. used 14000 text news documents (in addition to MATBN) for estimating the models' parameters in the paper, which was not released as a new corpus to enable replication of their work.

## 4.3 Strengths and Limitations

Our review has analysed speech summarisation methods and features based on the domain, content, and type of data. We surveyed more than 100 papers deploying a range of methods and architectures, which is useful for selecting the right methodology for a specific domain. A major strength of our review is that we used a systematic approach based on published guidelines [27] [126] to review five large repositories of research papers. We provide the detailed information in Electronic Supplement, including the definitions of methods and evaluation metrics, a brief description of publicly available corpora, the table of quantitative evaluation results and the table of information extracted from the reviewed papers. However, due to the search strategy, this review may have omitted papers not indexed by the databases we searched. This review may also not include recent works on abstractive summarisation and deep learning methods published after we conducted our search, due to the rapid pace of publication of machine learning research.

Our analysis highlighted several findings around the prevalent domains, architectures, and methods for summarisation. These are indicative of research trends in NLP and machine learning at large, which may guide the selection of summarisation methods. However, we also observed a substantial variability across the experimental setting of the analysed papers, in particular in terms of the corpora, training and test datasets used, features used summarisation, and importantly the evaluation metrics. Hence, the direct comparison of the obtained results and the performance analysis of the proposed methods in these papers is difficult.

## 4.4 Comparison to Prior Work

To the best of our knowledge, this is the first scoping review of automatic speech summarisation. There has been only one survey paper on automatic speech summarisation exploring different types of output summaries, features, methods, and evaluation metrics [24]. That work considered a smaller number of papers published before 2006 and only investigated a two-stage summarisation method containing key sentence extraction and sentence compaction. A follow-up review of spontaneous speech transcriptions was published by the same authors in 2008 and discussed speech corpora issues, speech recognition, acoustic models, speech structure extraction, and speech summarisation [25]. However, only a small part of the paper focused on speech summarisation, explaining different types of speech summarisation and reviewing methods for key sentence extraction and summary generation. More than a decade passed after the publication of these reviews, and since then new speech corpora in various domains and new methods of speech summarisation have been proposed, which are captured in our scoping review. Also, we analyse the surveyed papers according to a wider range of criteria, thus, painting a more encompassing picture of the speech summarisation research.

## 4.5 Open problems and challenges

Despite the progress to date in speech summarisation, several open problems still require significant effort. Whilst some of these problems are shared with text summarisation, many challenges are unique to the world of speech.

### 4.5.1 Speech Recognition Errors

Despite rapid advances in automatic speech recognition, these technologies still suffer problems that have a flow on effect when summarising speech content. Most summarisation studies assume that sentences or spoken units are correctly rendered following transcription [81], but this may not be the case for transcripts produced by ASR. ASR engines can have error rates of up to 40%. [65]. ASR error rates are lower for well-defined single-speaker tasks like interpreting TED talks but remain problematic, when tasks are less structured or involve multi-speakers e.g. audio from meetings. Machine learning, deep learning and language models can be used to correct ASR output errors, but limited work has been done in this area to date e.g. indexing techniques [4]. In general, language model-based methods such as ULM, RNNLM and their combinations are better able to handle of the effect of imperfect speech recognition on speech summarisation compared to graph-based methods. RNNLMs with syllable level units for example, can convert ASR generated words into overlapping syllables to form a vocabulary of syllable pairs for indexing [106]. In contrast, graph-based methods are particularly error-prone because of the poor performance of similarity measures like SM and ILP - used to compare pairs of sentences [106].

### 4.5.2 Speaker turn identification

Summarising speech involving multiple speakers remains challenging. Diarisation is the task of segmenting an audio stream into homogeneous units associated with different speakers. Deep learning techniques including those used by Google and Microsoft are equipped with a speaker diarisation capability. However, none of the ASR engines, including commercial ones, has solved the diarisation problem, completely [127, 128]. Accurately detecting sentences or spoken unit boundaries is a particular challenge, as incorrect unit boundary detection may lead to incorrect speaker identification across a sequence of sentences. Summarisation is also challenged when information is distributed across a sequence of sentences e.g. a question-answering sequence from multiple speakers, where answers may be short ("yes" or "no") and directly refer to the utterances of other speakers. As a result, the summarisation process might exclude either the question or the answer in a question-answer pair because of apparent low value. There have been exploratory efforts to determine the relevance of spoken units, speaker turn identification, and creating continuous links between cross-speaker information and question-answer pairs, for example based on speaker activity or interaction and dialogue acts [9, 17, 42, 112].

### 4.5.3 Speech disfluency

In any conversation, it is common to have different kinds of disfluency, including interruptions, overlapped speech, interleaved wrong starts (e.g. "I'll, let's talk about it"), filler phrases (e.g. "of course", "ok", "you know"), non-lexical filled pauses (e.g. "umm", "uh") [129] and redundancies. These disfluencies complicate the identification of the semantics content of speech and consequently can impede summarisation. Speech in broadcast news is the closest to structured text in having the lowest number of disfluencies, due to the professionally training of the presenters [125]. In meetings and interviews, disfluencies, filler phrases, redundancies and a lack of structure make them more challenging for summarisation. Summarising lectures with untrained speakers can also be problematic. Researchers have tried filler phrase detection [65] and MMR to remove redundant information as an extra criterion [82, 114], but with mixed success.

### 4.5.4 Direct speech summarisation

Speech summarisation typically follows a two-stage process of transcribing speech to text and then summarising the transcription. Generating a summary directly from speech without transcription is an alternate pathway that may side-step some fundamental problems in summarisation. A speech summary in this case would be an audio file that contains an extractive or abstractive concatenation of spoken words drawn from the original speech. Very few studies have explored the possibility of direct summarisation from speech [15, 16, 67], partly because speech to text is such a popular approach. Direct speech summarisation in the WordCloud study clustered recurrent patterns in speech [67]. Hidden Markov Models using acoustic and prosodic features [15] have also been applied to identifying repetitive speech patterns, as have a combination of computer vision techniques and a similarity matrix of spectral features [16]. Although performance of these systems is promising, there is space for new approaches. For example, once could allocate higher weights to the most frequent sound patterns using deep learning, computer vision techniques using spectrogram images or indeed, other audio signal representations.

### 4.5.5 Abstractive speech summarisation

Because of the complexity of natural language, abstractive summarisation is an open problem, even in the text domain, using deep learning [1, 130, 131]. However, the task is harder in the speech domain because of the higher possibility of error propagation in transcriptions. Abstractive approaches to summarisation have attracted more attention than extractive approaches. This in part is because of studies showing abstractive methods achieve superior ROUGE and readability scores compared to extractive approaches [23, 66], as well as research that shows users prefer abstract summaries over extracts [8]. Additionally, the need for real-time summarisation is critical for some applications e.g. a healthcare digital scribe that aims to listen to and summarise conversations on the fly. In such cases a summariser does not have access to the full final content of the conversation but may need to draw intermediate conclusions that require repair as more information becomes available as a conversation or speech progresses. It may be that RNNs and LSTM architectures along with language models will be of value in tackling these abstractive challenges, given their potential to discover semantic relationships between sequences of utterances needed to generate meaningful summaries [133].

# 5 Conclusion

We framed our scoping review of speech summarisation methods across four main domains (broadcast news, lectures, meetings and spontaneous conversation/interview) and also reviewed publicly available training corpora. As sentence classification using feature vectors is so prominent in the literature, we identified widely used features and discussed their performances. Methods in the review were also distinguished by output (text/speech), use of extractive or abstractive methods, and architecture.

Speech summarisation is a fast-growing field of research that has the potential to contribute to many application domains and tasks. At present however, the evidence for their effectiveness remains limited. The wide variety of approaches, tasks and study designs limits our ability to genuinely compare the effectiveness of much of the published research. For this reason, future research should report in a more standardised way, and use standard public corpora to assist with performance comparisons.

**ACKNOWLEDGMENTS**

This paper was funded by the National Health and Medical Research Council (NHMRC) grant APP1134919 (Centre for Research Excellence in Digital Health) and Programme Grant APP1054146.